# Large Language Models for Automating Structured Clinical Data Standardization: HL7 FHIR Use Case

## Álvaro Riquelme Tornel[1], Pedro Costa del Amo, PhD[2], and Catalina Martínez Costa, PhD[1]

[1]Departamento de Informática y Sistemas, Universidad de Murcia, CEIR Campus Mare Nostrum, IMIB-Pascual Parilla,

Murcia, Spain, [2]Latency, Spain

**ABSTRACT**

**Objective:** Our objective was to design and evaluate a semi-automatic pipeline leveraging large language models to transform structured clinical data into HL7 FHIR resources, while rigorously assessing accuracy, reliability, and security implications.
**Materials and Methods:** We applied the proposed methodology to the MIMIC-IV dataset, combining a set of embedding techniques, clustering algorithms, and a semantic retrieval generation technique. These components were integrated with diverse prompting strategies to guide GPT-4o and Llama 3.2 405b in mapping table attributes into standardized HL7 FHIR format.
**Results:** In the fully contextualized baseline scenario, our pipeline achieved 100% resource-level identification and an 95% confidence interval of (67.02%-73.88%) attribute level mapping over 119 attributes with GPT-4o, outperforming Llama 3.2 405b (43.79%-52.98%). Under the simulated real world scenario, GPT-4o shown a 95% confidence interval of (67.740% - 69.860%) tested in differents temperature configurations, while Llama 3.2 reached a confidence interval of (54.450% to 57.770%), This reliably demonstrates the feasibility of the semi-automated mapping using Large Language Models.
**Discussion:** These findings demonstrate that detailed, machine readable context (JSON schemas) and semantic clustering substantially narrow confidence intervals and reduce mapping ambiguity, while iterative prompting further refines outputs. Nonetheless, challenges remain, including incomplete source descriptions, occasional model hallucinations.
**Conclusion:** This study confirms the feasibility of LLM driven, context aware transformations of clinical data into HL7 FHIR, establishing a solid foundation for semi automating interoperability workflows. Future efforts will focus on fine tuning models with domain specific corpora, integrating controlled terminology services, extending support to additional standards (e.g., HL7 CDA, OMOP) and developing an interactive interface for expert validation and iterative improvement.

## INTRODUCTION

Efficient, standardized, and interoperable data exchange plays a crucial role in healthcare by supporting clinical practice and medical research. However, reusing clinical information presents several challenges, such as ensuring data security, integrating heterogeneous data from different systems, and maintaining data quality[1].

For decades, semantic interoperability standards have been proposed to enable accurate, meaningful, and consistent exchange of healthcare data across different systems and organizations. Examples are terminologies like SNOMED-CT[2], LOINC[3], etc. which define medical concepts (e.g. diseases, lab results, procedures, medications, etc.) and EHR standards and specifications such as HL7 FHIR[4], OMOP CDM[5], or openEHR[6] that reuse medical terminologies and define how to structure and exchange data across systems.

Organizing healthcare data according to a single international standard simplifies access, analysis, and integration across systems. However, in practice, data is often organized using locally developed information models, and even when standard models are adopted, multiple standards frequently coexist. Each organization may choose a different model, leading to a lack of semantic interoperability. Furthermore, even when a single standard is used, variations in its application can arise if implementation guidelines are inadequate or not consis-tently followed. Moreover, modeling data using these standards requires significant resources, expertise, and time, making implementation complex[7].

Large Language Models (LLMs) have emerged as powerful tools for automating and solving complex tasks with minimal effort. Their rapid evolution, with models like GPT-4o, Gemini, and Llama3, has expanded their applications across numerous fields. In healthcare, LLMs offer a promising solution to facilitate interoperability and data reuse[8].

This work presents a method for the semi-automatic standardization of clinical information by leveraging LLMs enhanced with specific domain knowledge through Retrieval Augmented Generation (RAG). The approach is implemented using the HL7 FHIR standard and the MIMIC IV structured dataset[9]. Furthermore, the study analyzes and compares the performance of different LLMs in supporting data standardization. It contributes to exploring the potential of LLMs in advancing semantic interoperability, reducing the need for resources such as domain experts and time.

## BACKGROUND AND SIGNIFICANCE

### HL7 FHIR for Data Interoperability

HL7 FHIR is an open standard designed to facilitate the exchange of healthcare data by defining reusable resources, such as *Medication*, *Observation*, or *Patient*, that capture common clinical entities[10]. These resources can be profiled and tailored for diverse clinical scenarios.

*Implementation Guides* (IGs) specify detailed conventions for the use of FHIR resources in specific contexts, thereby promoting consistency across heterogeneous systems. The broad adoption of FHIR is supported by its foundation on widely used web technologies, its RESTful architecture, which facilitates data exchange, and the extensive documentation available, which supports implementers in its application across use cases. Additionally, FHIR integrates medical terminologies such as ICD[11], SNOMED CT, and LOINC, facilitating the standardization of both structured and unstructured data[12].

When using FHIR for standardizing structured data, most approaches manually define the correspondences between the data and the corresponding FHIR representation. Then, data transformation is implemented through an ETL process[13–16]. For unstructured data, advanced natural language processing (NLP) techniques—provided by tools like *NLP2FHIR*—extract clinical entities from free-text documents and convert them into RDF-based FHIR resources[17].

All these approaches have in common the manual definition of correspondences from the data to the corresponding FHIR resources. Thus, it requires expertise in both the source data models and the FHIR specification, as well as significant effort to ensure semantic accuracy and consistency across resources.

### Large Language Models in Healthcare Interoperability

LLMs have become very useful tools for solving long-standing problems in managing and integrating clinical data. For example, scalable pipelines based on LLMs have been shown to improve the extraction of medical data, such as accurately identifying medication records, which increases the overall quality of available information.[18]. In addition, LLMs can transform structured laboratory data into natural language descriptions and then reconstruct it without losing significant information, map diagnostic codes between *ICD-9-CM* and *SNOMED-CT* with high accuracy, and extract medication names from unstructured discharge notes[19].

Building on these achievements, several studies have begun using LLMs for interoperability and data transformation tasks. For example, related works like[20] demonstrated that LLMs can generate FHIR bundles from patient reports, although they only studied a limited set of FHIR resources. Other work has used LLMs to automatically align medical intake forms with FHIR standards, reducing redundancy and manual effort[21]. In the healthcare field, combining LLMs with HL7 FHIR has opened new ways to process electronic health record (EHR) data more efficiently. For instance, *FHIR-GPT* converts clinical free text into structured FHIR resources, making data entry easier and improving interoperability. Nevertheless, only *MedicationStatement* resource was tested, limiting generalisability to other clinical resources.[22]. GPT-4 has also been used to clean and identify clinical concepts, supporting the mapping of these data to FHIR elements. However, this approach has not been exhaustively validated and requires further validation[23]. Moreover natural language interfaces powered by LLMs enable intuitive querying and summarization of EHRs[24].

However, important challenges remain. It is still difficult to handle rare or complex scenarios, ensure that all data are completely reliable, and maintain very strict privacy and security measures. Therefore, current research on private LLM deployments emphasizes the need for strong fine-tuning and local solutions, showing that LLM applications in healthcare continue to evolve[25]

The integration of HL7 FHIR and LLMs represents a significant advancement in processing healthcare data. By combining robust structured data models with advanced natural language processing and external data retrieval, these technologies worked together to improve system compatibility, streamline data integration, and enhance data interoperability in healthcare.

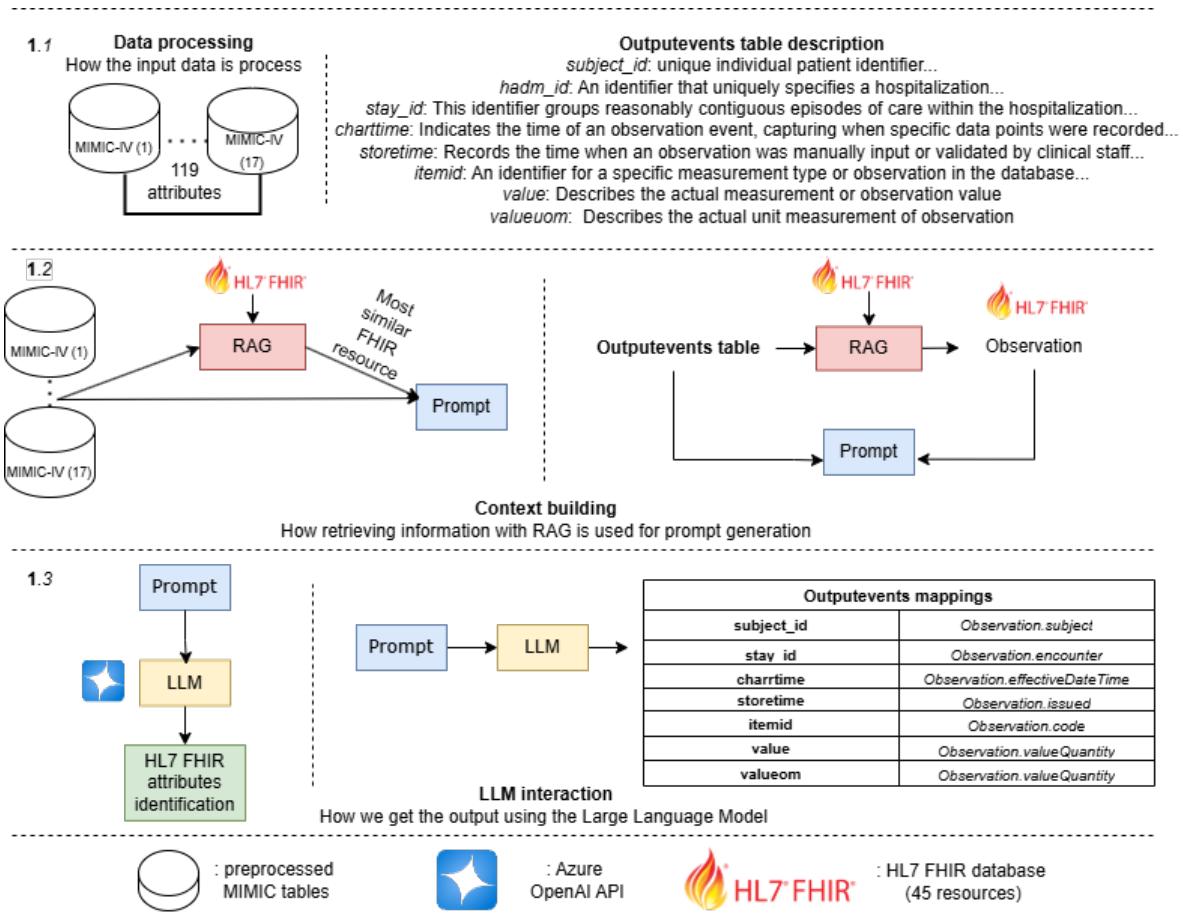

**Figure 1.** Figure illustrating the designed pipeline and a data example for the baseline scenario. The pipeline is divided into three steps. The transformation and the results obtained are depicted as the stages progress (from top to bottom). This example is presented for the Outputevents table, one of the 17 tables in MIMIC-IV.

## METHODS

The proposed methodology for the semi-automatic standardization of clinical information consists of three main steps: **Data Processing**-preprocessing the input data and transforming it into a more easily understood format; **Context Building**-using the preprocessed data to extract as much relevant information as possible to describe the task to the model; **LLM Interaction**-having the model execute the requested task. We have applied it to HL7 FHIR in two scenarios, to assess the methodology in different contexts. The first scenario served as a baseline, using initial data that were encapsulated and well contextualized. In contrast, the second scenario simulated a real data provider, where the data source lacked both contextualization and encapsulation.

For the evaluation we used the HL7 FHIR representation of both the MIMIC-IV and MIMIC-IV Emergency Department (MIMIC-IV-ED) databases, which contain de-identified health data from intensive care unit (ICU) patients at Beth Israel Deaconess Medical Center between 2008 and 2019. These datasets include comprehensive patient information, including laboratory measurements, medications, and vital signs, supporting a wide range of healthcare research. Its HL7 FHIR representation consists of 24 profiles for MIMIC-IV and 6 profiles for MIMIC-IV-ED, including coded value sets.

As performance metric we defined, resource-level identification as the proportion of tabular data correctly matched to their corresponding HL7 FHIR resource and attribute-level mapping accuracy as the proportion of tabular attributes correctly matched to their corresponding HL7 FHIR elements.

The following sections provide a detailed description of each step of the methodology, as applied across the two evaluated scenarios.

### Baseline Scenario

In the baseline scenario (see Figure 1), we evaluated the methodology within a simplified setting.

In the *Data Processing step* we used the MIMIC-IV dataset and selected 17 tables comprising 183 attributes,

following we mapped those attributes manually to FHIR standard. Each selected table included a description of its intended use case, along with detailed metadata for all contained attributes.

Following this selection, during the *Context Building step*, we extracted contextual information by generating embeddings for each table. Using these embeddings, we identified the most closely corresponding HL7 FHIR resource from a set of 45 possible candidates. Finally, in the *LLM Interaction step*, this information was used to guide interactions with a LLM, facilitating the mapping of individual attributes. We explored multiple prompt formulations and interaction strategies to assess their effectiveness across different scenarios. In the following, the application of the three methodological steps is detailed.

**Data Processing**

Given the size and complexity of the original dataset, we performed a preprocessing phase which involved an exhaustive review of the initial 183 attributes to evaluate their equivalence or feasible representation within the HL7 FHIR standard. During this analysis, we identified and excluded redundant attributes and those lacking a feasible FHIR representation, either due to semantic misalignment with the standard. As a result, we selected a refined subset of 119 candidate attributes that could be transformed to FHIR effectively.

Figure 1 illustrates this process using one of the selected tables, named *Outputevents*, shown in step 1.1. It shows the data description of this table. Afterwards, the table was converted to JSON format to facilitate data management and understanding.

**Context Building**

In this step, we built several corpora to support the identification of the most appropriate HL7 FHIR resources for representing the source data using semantic similarity and embeddings. For each MIMIC table, we created a corpus consisting of the descriptions, their associated use cases, along with detailed metadata for all contained attributes. Then, we built another corpus with the descriptions of the 45 HL7 FHIR resources extracted from the official standard documentation. We aimed to identify the HL7 FHIR resource most similar to each attribute grouping, since the data were contextualized and perfectly encapsulated.

To assess semantic similarity, we applied multiple embedding models and measured the cosine similarity[26] between each MIMIC IV table's description and the HL7 FHIR resource corpus. The initial approach, which relied only on word-frequency embeddings, did not produce satisfactory results. Therefore, we adopted a hybrid retrieval strategy that combined several complementary embedding models: *TFIDF*[27], *BM25*[28], *Universal Sentence Encoder*[29], and *Word2Vec*[30].

For each model, we computed similarity rankings between the source tables and FHIR resources. These individual rankings were then aggregated using the *Reciprocal Rank Fusion* (RRF) algorithm[31], resulting in a global similarity ranking from which the most appropriate FHIR resource was selected for each table.

This hybrid methodology proved highly effective, achieving 100% accuracy in ranking the correct HL7 FHIR resource as the top similar for each evaluated attribute grouping. Figure 1, specifically step 1.2, illustrates the Retrieval Augmented Generation process. As an example, the *outputevents* table was matched to the HL7 FHIR resource *Observation* based on its description and metadata. This step effectively established the contextual foundation necessary for subsequent prompt-based interactions.

**LLM interaction**

After identifying the corresponding HL7 FHIR resource(s) for each table, each prompt combined this contextual information with detailed information about the source table, clearly instructing the LLM to map each column to the appropriate HL7 FHIR attribute. Instructions clarified that a single column might correspond to multiple FHIR attributes and emphasized the need to consider both column names and representative values. The LLM was also directed to validate its mappings carefully before producing a final output in a predefined JSON format linking each column to its corresponding FHIR attributes.

Based on the above, we experimented with several prompt strategies to optimize mapping accuracy. Specifically, we evaluated four custom methods: *Self-Reflexive*, *Mixture of Prompts (MoP)*, *5 Serial Schema*, and *5 Serial NoSchema*. The *Self-Reflexive* method required the model to first analyze its initial response internally and then, through another request, refine or correct its previous answer. The *Mixture of Prompts (MoP)* approach involved using several distinct prompts to reduce potential biases. Additionally, the *5 Serial Schema* strategy provided a structured sequence of five prompts, each including a JSON schema for clearer guidance, whereas the *5 Serial NoSchema* approach used a similar five-step sequence without the JSON schema, allowing us to evaluate the model's performance without structured contextual help.

The parameter values were selected as *temperature = 0*, aiming to minimize the randomness in the model's responses. Additionally, the parameter *functions* provided the model with HL7 FHIR resources represented as JSON schemas obtained from the hybrid retrieval system methodology, enabling structured and precise responses. By setting *function_call="auto"*, the model automatically decided when and which predefined functions to invoke, ensuring that results adhered to the structured format specified in the prompt instructions. As a result, each table attribute was successfully mapped to its corresponding HL7 FHIR attribute, as illustrated in Figure 1, step 1.3.

A key component of this process was the implementation of the *structured_output* functionality[32], which enforced a strict response format. This feature offered two main advantages: (1) it grounded the LLM in domain-consistent context derived from the selected HL7 FHIR resource, and (2) it ensured adherence to a precise JSON structure for mapping outputs. This greatly reduced the likelihood of generating unstructured or irrelevant responses, enhancing the reliability of the final mappings.

### Real World Scenario

The second scenario simulates a more realistic setting in which data is provided within a single table reflecting how clinical datasets are often encountered in practice. To construct this scenario, based on MIMIC IV data, we built a table comprising 68 columns / attributes, with the column order randomized to eliminate structural cues. Only column descriptions and representative values were made available, mimicking the typical lack of schema metadata in real-world data integration tasks.

Then, in the *Data Preprocessing step* we applied unsupervised clustering to group related attributes and to better understand their context. During the *Context Building step*, we applied different embedding models to each cluster and using semantic similarity with the HL7 FHIR resource corpus, we identified the five HL7 FHIR resources most semantically aligned with each cluster.

Finally, in the *LLM Interaction step*, using this contextual information, both the cluster groupings and the top-matching FHIR resources, we constructed refined prompts for LLM interaction. These prompts were designed based on insights gained from the earlier baseline scenario and tailored to guide the model in mapping individual attributes within each cluster.

**Data Processing**

From the original MIMIC-IV dataset, a subset of 68 attributes was selected and consolidated into a single table with randomly arranged columns. This selection had been made previously by MIMIC developers, who then mapped it to FHIR. This reduction was intended to establish a validation subset for evaluating the accuracy of the attribute mappings. The dataset of 45 HL7 FHIR resources, including their descriptions, remained unchanged.

Initial experiments showed that applying similarity at the level of individual attributes did not reliably identify the correct HL7 FHIR resources. To address this limitation, we used unsupervised clustering to group semantically related attributes, thereby enriching their contextual representation within the embedding space and enhancing retrieval performance.

We evaluated several embedding models to vectorize the attribute descriptions. In addition, we applied several clustering algorithms capable of grouping attributes based on their semantic similarity: *KMeans*[33], *Agglomerative Clustering*[34], *DBSCAN*[35], *BIRCH*[36], *OPTICS*[37], and *Spectral Clustering*[38].

To identify the optimal clustering method, we assessed their performance using standard clustering quality metrics, selecting the algorithm and configuration with the highest scores: *Silhouette*[39], *Calinski-Harabasz*[40], and *Davies-Bouldin*[41].

As illustrated in Figure 2 (step 2.1), the unified table was processed to generate semantically meaningful clusters. These were labeled as *Cluster X*, each containing a non supervised group of attributes along with their respective names and descriptions. For example, *Cluster 3* is shown, demonstrating the output of the clustering process.

**Context Building**

After *Data Preprocessing*, we applied the same embedding models as in the baseline scenario to compute similarities between each cluster (attribute grouping) and the FHIR resources. However, instead of selecting only the single most similar resource for each grouping, we retained the top five most similar resources—since the groupings are unsupervised and several FHIR resources can be very alike. These initial embeddings achieved an accuracy of 87.9% in correctly identifying HL7 FHIR resources from individual table attributes. To further improve both clustering coherence and resource identification accuracy, we explored embedding models specifically trained on biomedical and clinical texts: *pubmedbert-base-embeddings*[42], *MedEmbed-large-v0.1*[43], *ClinicalBERT*[44], and *biobert-v1.1*[45].

For each attribute within the cluster, the top five most similar HL7 FHIR resources were identified based on cosine similarity. The use of specialized embeddings led to a significant increase in accuracy, achieving a 94% success rate in correctly identifying the relevant FHIR resources.

Once the most relevant HL7 FHIR resources were selected for each cluster, we proceeded with LLM-based interaction to map individual attributes from the input table to the attributes defined in the selected FHIR resources.

Figure 2, step 2.2, illustrates this process using Cluster 3 as an example. At this stage, only the attribute descriptions were provided to the similarity retrieval system since no table-level context was available.

**LLM interaction**

Compared to the baseline scenario, we build an enhanced version of the prompt by emphasizing the importance of internal reasoning to ensure accurate and semantically appropriate matches, while suppressing the display of intermediate reasoning steps. The prompt also referenced the official HL7 FHIR documentation as the authoritative source, reinforcing the need for precision and standards compliance.

The model was instructed to identify the three most relevant HL7 FHIR attributes for each input attribute. As in the baseline case, responses were required to follow a strict JSON structure. In cases where fewer than three suitable matches were found, placeholder values were inserted to maintain consistent formatting.

To ensure deterministic behavior, the parameters *temperature=0* and *top_p=0* were applied, minimizing randomness and constraining the model to select only the most confident predictions. This combination of settings effectively eliminated variability, producing highly consistent outputs.

The *functions* parameter defined a set of callable tools representing HL7 FHIR resources as JSON schemas—generated previously through the similarity retrieval system. These allowed the model to internally call these functions when necessary, generating structured and targeted responses. Finally, with the parameter *function_call="auto"*, the model was granted autonomy to invoke these functions as needed, selecting the most appropriate schema based on the task context. Figure 2 (step 2.3) illustrates the final mapping process. As shown, each attribute within a cluster was successfully matched to up to three candidate HL7 FHIR attributes.

## RESULTS

Completing the baseline scenario allowed us to identify the most effective prompt for each language model. Tables 1 and 2 report the 95 % confidence intervals for **GPT-4o** and **Llama 3.2 405b**, respectively. For **GPT-4o**, the *Reflexive Prompt* achieved the highest performance range (**67.02–73.88**), followed by *MoP* (**64.50–70.89**), *Reflexive Serial Schema 5* (**60.20–67.13**) and *Reflexive Serial No Schema 5* (**61.73–62.10**). In contrast, for **LLaMA 3.2 405b**, the *MoP* approach yielded the best performance (**43.79–52.98**), followed by the *Reflexive Prompt* (**40.28–53.49**), *Reflexive Serial No Schema 5* (**28.85–49.15**), and *Reflexive Serial Schema 5* (**28.49–35.62**). From these outcomes, the highest-performing prompt for each model was selected for deeper analysis in the second scenario.

In the second scenario, the selected prompt was tested under three temperature settings (t = 0, 0.5 and 1) to measure how randomness affects accuracy. Table 3 shows that **GPT-4o** achieved its highest mean accuracy at **t = 0.5** (**68.8%**, 95% CI: 67.7–69.9), with slightly lower results at t = 0 (**68.2%**, CI: 66.8–69.6) and t = 1 (**67.8%**, CI: 65.3–70.2). Table 4 shows that **LLaMA 3.2 405b** performed best at **t = 0** (**56.1%**, CI: 54.5–57.8), but performance declined at t = 0.5 (**52.2%**, CI: 48.8–55.7) and t = 1 (**51.6%**, CI: 48.5–54.7).

Overall, **GPT-4o** consistently outperformed **LLaMA 3.2 405b** across all conditions and demonstrated greater stability, as reflected by its narrower confidence intervals.

Additionally, the consistently narrow confidence intervals observed across both experiments indicate low variability in the models' outputs, reinforcing the reliability and reproducibility of their performance.

| *Baseline Scenario – GPT 4o* | |
|---|---|
| **Prompts** | **Range** |
| ***Reflexive Prompt*** | **[67,02–73,88]** |
| *MoP* | [64,50–70,89] |
| *Reflexive Serial Schema 5* | [60,20–67,13] |
| *Reflexive Serial No Schema 5* | [61,73–62,10] |

**Table 1.** This table summarizes the 95 % confidence intervals obtained for each prompt strategies evaluated with **GPT-4o** in the Baseline Scenario (N = 4).

| *Baseline Scenario – Llama 3.2 405b* | |
|---|---|
| **Prompts** | **Range** |
| *Reflexive Prompt* | [40,28–53,49] |
| ***MoP*** | **[43,79–52,98]** |
| *Reflexive Serial Schema 5* | [28,49–35,62] |
| *Reflexive Serial No Schema 5* | [28,85–49,15] |

**Table 2.** This table summarizes the 95 % confidence intervals obtained for each prompt strategies evaluated with **Llama 3.2 405b** in the Baseline Scenario (N = 4).

| Real World Scenario GPT-4o | | |
|---|---|---|
| t=0 | t=0.5 | t=1 |
| 66.780 to 69.630 | 67.740 to 69.860 | 65.330 to 70.180 |

**Table 3.** This table summarizes the 95 % confidence intervals obtained for the refined prompt when evaluated with **GPT 4o** in a Real World Scenario (N = 10).

## DISCUSSION

In this work, we propose and evaluate a semi-automatic methodology to map tabular data to the HL7 FHIR interoperability standard using Large Language Models (LLMs). Our methodology integrates Retrieval-

| Real World Scenario for Llama 3.2 405b | | |
|---|---|---|
| t=0 | t=0.5 | t=1 |
| 54.450 to 57.770 | 48.770 to 55.700 | 48.530 to 54.740 |

**Table 4.** This table summarizes the 95 % confidence intervals obtained for the refined prompt when evaluated with **Llama 3.2 405b** in a Real World Scenario (N = 10).

Augmented Generation, Prompt Engineering, and Semantic Clustering to enhance both accuracy and consistency in data transformation. We tested two settings. In the Baseline Scenario, *GPT-4o* significantly outperformed *Llama 3.2 405b*: for the Reflexive Prompt method, GPT-4o achieved a 95 % confidence interval of 67.02–73.88 (width 6.86), whereas Llama 3.2 managed only 43.79–52.98 (width 9.19) with (Mixture of Prompts) MoP. In the Real-World Scenario, GPT-4o's performance remained stable across different temperature settings, while Llama 3.2 showed greater variability—highlighting GPT-4o's inherent robustness under deterministic and near-deterministic configurations. Moreover, providing detailed JSON schemas narrowed confidence-interval widths, demonstrating that clear, machine-readable context is essential for reliable mappings. Semantic clustering grouped related attributes, enriching the model's context and guiding it toward more relevant FHIR resource subsets. Nonetheless, important limitations persist. Missing source-data descriptions sometimes led to inconsistent mappings, and we observed isolated "hallucinations" where plausible but incorrect mappings were proposed. These issues underscore the need for manual validation workflows. Future work will fine-tune LLMs on FHIR-specific corpora, extend support to standards such as OMOP, OpenEHR or HL7 CDA, and integrate unstructured clinical narratives to assess performance in mixed-data contexts. We also plan to develop an interactive UI for mapping visualization and expert feedback, and to benchmark open-source LLM variants to improve accessibility and customization in real-world clinical environments. Together, these advances will drive more automated, accurate, and reliable healthcare data integration.

## CONCLUSION

This study confirms the feasibility of employing Large Language Models (LLMs) for clinical data mapping tasks, underscoring their substantial potential while also identifying key areas for future improvement. By integrating structured contextual information, iterative prompting strategies, and semantic clustering techniques, a robust and scalable methodology is established. Furthermore, this research provides a scientific foundation for transforming clinical data via interoperability standards, laying the groundwork for automating or semi-automating traditionally complex data mapping tasks through advanced LLMs. These innovations offer clinical data modeling experts promising avenues toward enhanced efficiency and effectiveness in their workflows.

## ETHICS STATEMENTS

## FUNDING

This work was supported by the grant RYC2020-030190-I, in part by the Horizon Europe HORIZON-HLTH-2021-TOOL-06-03 under Grant 101057603, and in part by Horizon-HLTH-2022-Tool-12-Two-Stage under Grant 101080875.

## CONFLICT OF INTEREST

## DATA AVAILABILITY

The dataset used in this study is not publicly accessible; obtaining access requires completing a specific training course provided by PhysioNet. However, a publicly available demonstration dataset can be accessed for reference and experimentation purposes at https://physionet.org/content/mimic-iv-demo/2.2/. The various prompts and complete results of the data transformations are available on GitHub at https://github.com/alvumu/dataInteroperabilityLLM.git

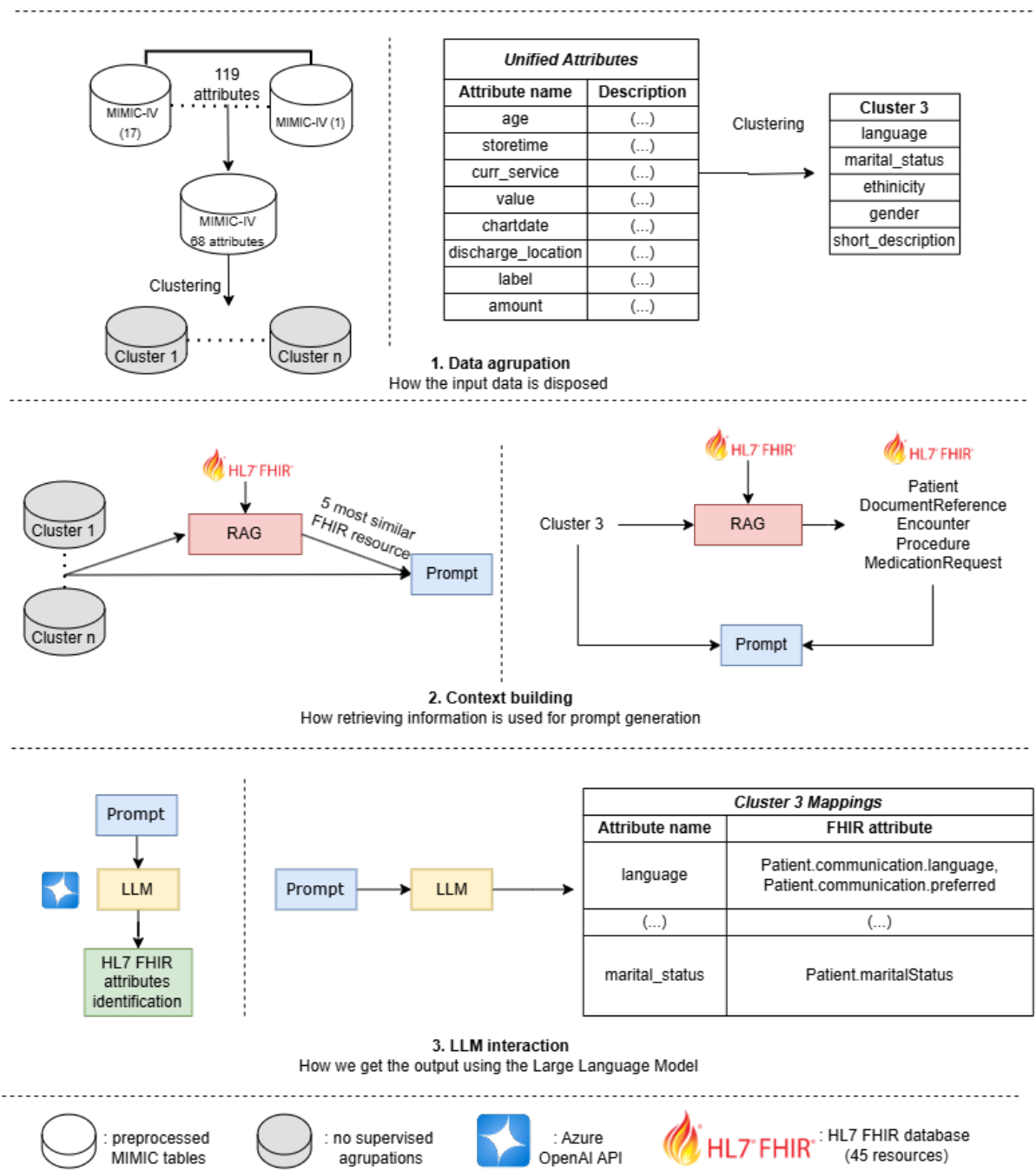

**Figure 2.** Figure illustrating the designed pipeline and a data example for the real world scenario. The pipeline is divided into three steps. The transformation and the results obtained are depicted as the stages progress (from top to bottom). This example is presented using the unified table in a random arrangement and illustrates one of the unsupervised groupings (Cluster 3).